\pdfoutput=1

\documentclass[10pt,twocolumn,letterpaper]{article}

\usepackage[pagenumbers]{cvpr} 

\usepackage{graphicx}
\usepackage{amsmath}
\usepackage{amssymb}
\usepackage{booktabs}
\usepackage{multirow}

\usepackage[pagebackref,breaklinks,colorlinks]{hyperref}

\usepackage[capitalize]{cleveref}
\crefname{section}{Sec.}{Secs.}
\Crefname{section}{Section}{Sections}
\Crefname{table}{Table}{Tables}
\crefname{table}{Tab.}{Tabs.}

\def\cvprPaperID{27} 
\def\confName{CVPR}
\def\confYear{2023}

\begin{document}

\title{Mutual Exclusive Modulator for Long-Tailed Recognition}

\author{Haixu Long$^{1}$, Xiaolin Zhang\thanks{Corresponding author.}, Yanbin Liu$^{2}$, Zongtai Luo$^{3}$, Jianbo Liu$^{3}$\\
$^{1}$University of Science and Technology of China, \\
$^{2}$Australian National University \\
$^{3}$SenseTime Research\\
{\tt\small hxlong@mail.ustc.edu.cn, solli.zhang@gmail.com, 	csyanbin@gmail.com} \\
{\tt\small luozongtai@hotmail.com, liujianbo@link.cuhk.edu.hk}
}
\maketitle

\begin{abstract}
   The long-tailed recognition (LTR) is the task of learning high-performance classifiers given extremely imbalanced training samples between categories.
Most of the existing works address the problem  by either enhancing the features of tail classes or re-balancing the classifiers to reduce the inductive bias. 
In this paper, we try to look into the root cause of the LTR task,~\ie, training samples for each class are greatly imbalanced, and propose a straightforward solution.
We split the categories into three groups,~\ie, many, medium and few,  according to the number of training images.
The three groups of categories are separately predicted to reduce the difficulty for classification.
This idea naturally arises a new problem of \textit{how to assign a given sample to the right class groups?}
We introduce a mutual exclusive modulator which can estimate the probability of an image belonging to each group.
Particularly, the modulator consists of a light-weight module and learned with a mutual exclusive  objective.
Hence, the output probabilities of the modulator encode the data volume clues of the training dataset. They are further utilized as prior information to guide the prediction of the classifier. 
We conduct extensive experiments on multiple datasets,~\eg, ImageNet-LT, Place-LT and iNaturalist 2018 to evaluate the proposed approach.
Our method achieves competitive performance compared to the state-of-the-art benchmarks.
\end{abstract}

\section{Introduction}
\label{sec:intro}

\begin{figure*}
  \centering
  \begin{subfigure}{0.44\linewidth}
    \centering
    \includegraphics[width=0.85\linewidth]{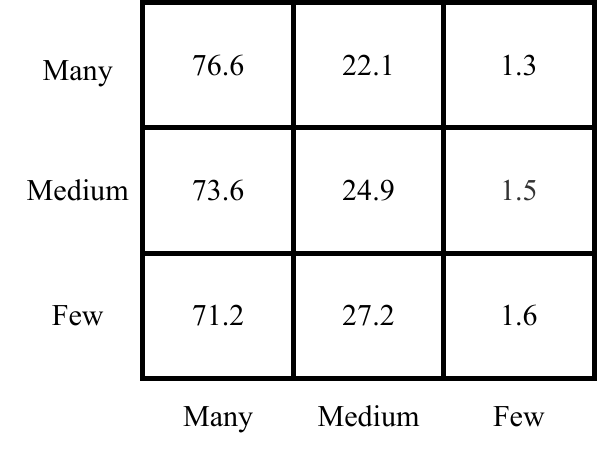}
    \caption{Misclassification percentage between groups.}
    \label{fig:motivation_a}
  \end{subfigure}
  \hfill
  \begin{subfigure}{0.48\linewidth}
    \centering
    \includegraphics[width=0.85\linewidth]{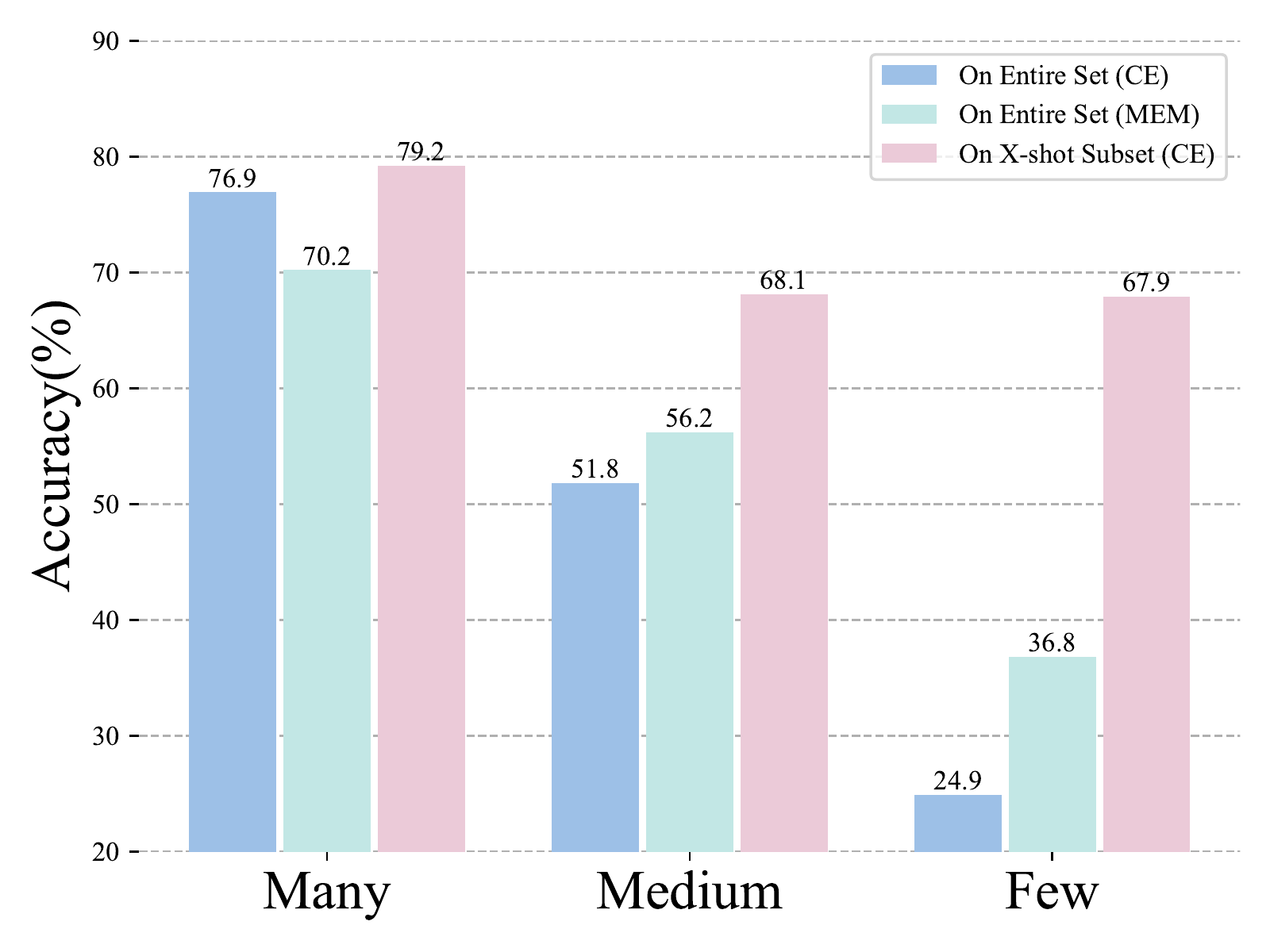}
    \caption{Comparision on ImageNet-LT.}
    \label{fig:motivation_b}
  \end{subfigure}
  \caption{(a) The digits indicate the percentage of images being misclassified to the wrong groups. For example, the digit $d_{i,j}$ on $i$ row, $j$ column means the misclassified images in group $i$ has $d_{i,j}$ percent be misclassified to group $j$. Analysis is conducted on ImageNet-LT dataset with Vit-Base. (b) Evaluation accuracy compares between the entire set and the $X$-shot subset, where $X \in \{\text{Many, Medium, Few}\}$. For the entire set evaluation, which is also the common evaluation approach, consider all classes in this process. However, for $X$-shot evaluation, we only use the $X$-shot classes output rather than overall classes to calculate the $X$-shot accuracy..}
  \label{fig:motivation}
\end{figure*}

In the past years,  deep learning has achieved significant progress in computer vision~\cite{krizhevsky2012imagenet}.
The huge success in deep technologies is inseparable from the availability of high-quality large-scale datasets,~\eg, ILSVRC~\cite{russakovsky2015imagenet}, MS COCO~\cite{lin2014microsoft}, and Places~\cite{zhou2017places}.
In contrast with these canonical datasets which are manually well-balanced across different categories \emph{w.r.t} training data samples, the real-world data are always extremely skewed and exhibit long-tailed distribution. Most of the samples are congregated on a few of categories,~\ie, head classes, while the rest categories,~\ie, tail classes, possess very limited samples.
Traditional models learned on such datasets perform very weak generalization ability  and obtain poor recognition accuracy on tail classes.

To alleviate the skewness and increasing the performance of tail classes, previous works can be roughly summarized into two streams.
The first stream aims to increase the representation capability of models and improve the extracted features for tail classes by seeking various constraints and network architectures~\cite{liu2019large, cui2019class, cui2021parametric, yang2020rethinking}.
Another line of works focuses on the adjustment of the decision boundaries of classifiers~\cite{he2009learning,shen2016relay,mahajan2018exploring,cui2019class}. For example, \cite{kang2019decoupling} proposes to boost the recognition performance of tail classes by adjusting the cardinality of the  classifier layer so as to adjust the separating hyperplane between categories.

In this paper, we first analyze the misclassified samples across different classes. For simplicity, we follow the same setting in~\cite{liu2019large} and divide the given dataset into three groups, \emph{i.e.}, \emph{Many}, \emph{Medium} and \emph{Few}, on the basis of the samples in each classes.
Figure~\ref{fig:motivation_a} depicts the proportion of misclassified images from one group to another.
For the misclassified images in ``Medium'' group, 73.6\% of them are predicted into the ``Many'' group.
Similarly, for ``Few'' group, 
71.2\% of the wrongly predicted images are attributed into the ``Many'' group and 27.2\% are assigned into the ``Medium'' group.
Such an observation naturally arises a question: \textit{Can reducing the misclassification rate between groups facilitate the accurate prediction?}
To illustrate this, we hypothesize an ideal case that the ground-truth group that each sample belongs to is available. We then compute the precision on each group separately and plot it in the light pink column in Figure~\ref{fig:motivation_b}. Compared to the case without group prior (light blue column), the dramatically improved performance of each group, for example, 67.9 \emph{v.s.} 24.9 in ``Few'' group, answers yes to this question.

We introduce a Mutual Exclusive Modulator (MEM) to realize this purpose.
MEM aims to learn and predict the mutually exclusive guidance of the groups for a given image.
This prior-guidance assigns a set of selection probability on the three groups and enables the classifier to predict the right category 
within the chosen group.
With the guide of MEM, the classifier can be re-adjusted and achieve a more accurate prediction from the guided label space.
To learn the MEM guidance, we follow the same idea of the network activations,~\ie, the output of MEM is activated if the group of an image is rightly predicted, otherwise it is depressed.
Particularly, an embedding vector is employed to represent the group a given image belongs to. 
The magnitude of the embedding is optimized to be large~\wrt the correct groups.
Otherwise, the magnitude should be small and close to zero for the wrong groups.
To achieve this goal of the feature magnitudes, we propose an objective function to learn the data-aware embedding in the training process.
In detail, we first train a standard classification model by supervised learning. Based on the representation of the standard model, we train a mutual exclusive modulator to estimate the activation values for each group.


After harvesting the group guidance information, it is further utilized to obtain precise category predictions. We propose a data-aware classifier by fusing the group information into the estimation of the accurate labels in a soft manner. In particular, the data-aware classifier has the analogical property with the canonical classifier, so it can also be trained by backward propagation directly. In Figure~\ref{fig:motivation_b}, we reveal that the proposed method increases the recall rate to facilitate the classification precision in each group (the light green column).
Especially, the accuracy in the few group significantly improves by 11.9\%.
Our contribution can be summarized as follows:
\begin{itemize}
\item We analyze the relationship between the categories and the number of training samples by splitting them into different groups, and explore that the causes of low accuracy for tail classes are the low recall rates;
    \item We propose a novel mutual exclusive modulator to boost the recall of tail classes so as to improve the overall accuracy of long-tailed recognition;
    \item We conduct extensive experiments compared with the most relevant decoupled learning methods,~\ie, $\tau$-normalized, cRT, and LWS, and the state-of-the-art for long-tailed recognition.
    We show that our method achieves a more balanced performance between head and tail classes and outperforms the state-of-the-art by a nontrivial margin.
\end{itemize}

\section{Related Work}
\label{sec:related work}

\noindent\textbf{Long-Tailed visual recognition.} In a general way, methodologies in long-tailed recognition can be categorized as classes re-balancing, multi-stage learning, and ensemble learning. Classes re-balancing can be further divided into two types: re-sampling and re-weighting. For re-sampling, The most intuitive approach is under-sampling head classes or over-sampling tail classes to achieve more balance between head classes and tail classes \cite{he2009learning, han2005borderline}. Another way for re-sampling is to apply a class-balanced sampling based on the cardinality of each class \cite{shen2016relay, mahajan2018exploring}. For re-weighting, various methods propose to assign different losses to different instances \cite{lin2017focal, ren2020balanced, cui2019class}. Multi-stage based methods deal with long-tailed datasets by conducting a multi-steps schema~\cite{kang2019decoupling, li2021self}. For instance, \cite{kang2019decoupling} decouples the learning procedure into representation learning and classifier learning. More recently, ensemble-based methods usually adopt multiple experts to reduce the model variance, \eg, RIDE \cite{wang2020long}, LFME \cite{xiang2020learning}, TADE \cite{zhang2021test}, BBN \cite{zhou2020bbn}, and NCL \cite{li2022nested}. For example, LFME constructs three experts in the cardinality-adjacent subset to reduce the imbalance in training with the assumption that training on these subsets is better than jointly trained counterparts. 

\noindent\textbf{Self-supervised pretraining.} Self-supervised learning unleashes the potential of vision transformers \cite{caron2021emerging,chen2021empirical,he2022masked}. In the past few years, contrastive learning is very popular, which aims to learn invariances from different augmented views of images \cite{he2020momentum, chen2020simple, grill2020bootstrap}. Recently, Masked Image Modeling (MIM) \cite{xie2022simmim,liu2022mixmim,he2022masked,gao2022convmae} become more and more prevalent for vision transformers. MIM is the task that reconstructs image content from a masked image. Mask Autoencoders (MAE) \cite{he2022masked} is a recent representative work. MAE builds an asymmetric encoder and decoder to reconstruct the corrupted input images in which most tokens are randomly masked.

\noindent\textbf{Out-of-distribution detection.} Out-of-distribution detection \cite{hendrycks2016baseline,liang2017enhancing,devries2018learning,bendale2016towards,lakshminarayanan2017simple} is crucial to ensuring the reliability of the learning system \cite{yang2021generalized}. It is a task to discriminate whether a sample in the inference is from a different distribution of the training data. Some approaches reject the out-of-distribution samples by setting a threshold on maximum softmax scores, they assume that the out-of-distribution samples will have a low maximum softmax score. However, different from the methods making a such assumption on out-of-distribution samples, \cite{dhamija2018reducing} proposes two simple yet effective loss functions, the Entropic Open-Set loss, and Objectosphere loss, to jointly train the model with in-distribution data and out-of-distribution data. The Objectosphere loss attempts to increase the feature magnitude for in-distribution data and decrease it for other data. Motivated by \cite{dhamija2018reducing}, we design a regularization objective for the sub-network of MEM to activate the positive samples and depress the negative samples.


\section{Method}
\label{sec: method}

\begin{figure*}[h]
\begin{center}
\includegraphics[width=1.0\linewidth]{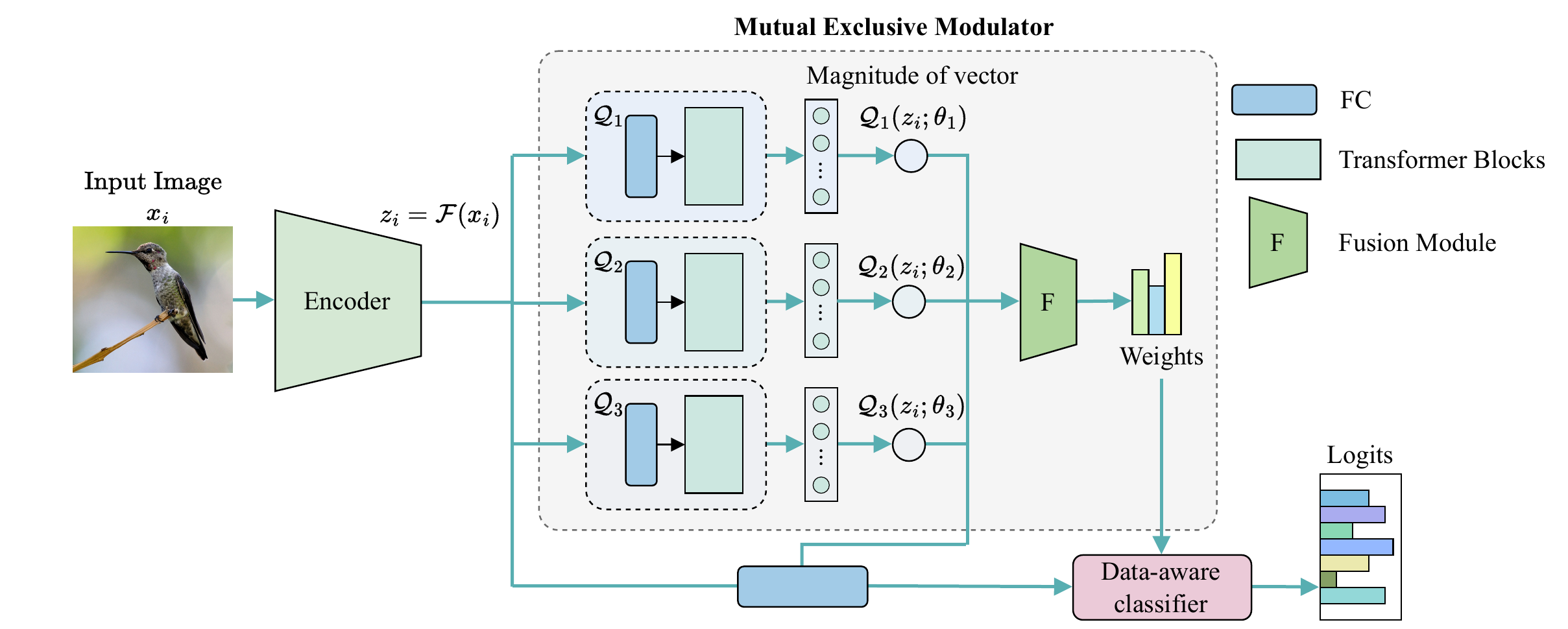}
\end{center}
\caption{An illustration of our proposed Mutual Exclusive Modulator (MEM). MEM consists of three sub-networks and a fusion module. The sub-network maps the representation $z_i$ to a latent vector that implies the signal of the group. To generate the adaptive weights, the fusion module encodes the magnitude from each sub-network and top K logits of each group from classifier.}
\label{method}
\end{figure*}

For long-tailed recognition, various approaches of re-sampling~\cite{shen2016relay,mahajan2018exploring} and re-weighting~\cite{cui2019class,ren2020balanced} aim to alleviate the bias towards the head classes and most of them adhere to joint learning representation and classification.
\cite{kang2019decoupling} decouple the learning procedure into representation learning and classifier learning, which presents a great potential way from a different perspective. 
We follow this schema and propose a novel Mutual Exclusive Modulator (MEM) to receive more accurate predictions. 
Practically, we first train a standard classification model by supervised learning. 
Then, based on the representation of the standard model, we train an individual mutual exclusive modulator to generate a group of adaptive weights. 
By encoding the adaptive weights, the data-aware classifier act on logits in a soft-routing manner to boost the classification accuracy.

\subsection{Mutual exclusive modulator}

Motivated by the observation in Figure~\ref{fig:motivation_a} that most of the misclassified samples  are incorrectly recognized into the wrong group rather than the group they belong to. 
The performance can be significantly promoted if the samples were recalled in the right group, as illustrated in Figure~\ref{fig:motivation_b}. 
Thus, we explore a new way of trying to recall the misclassified samples to the right group and propose a novel module~\ie, Mutual Exclusive Modulator (MEM). Without loss of generality, we split the classes into three mutual exclusive groups and denote them as $\mathcal{G}=\{\mathcal G_1, \mathcal G_2, \mathcal G_3\}$, following the previous practices~\cite{liu2019large}. The \textit{mutual exclusive} indicates that a category can only be assigned into one specific group according to the number of training images. 

Figure~\ref{method} depicts the overall structure of our approach. Given an image $x_i$, the encoder $\mathcal F$ transfers $x_i$ to representation $z_i$, $z_i=\mathcal F(x_i)$. With input $z_i$, the MEM generates a group of adaptive weights via two procedures. First, the sub-networks inside the MEM map the representation $z_i$ to latent vectors which imply the signal of the groups. To explicitly express the signal, we adopt the magnitude (L2-norm) of the latent vector to represent it. Such a magnitude represents the influence contributed by each group. In order to learn it, we design an objective to regularize the sub-networks. Afterward, we generate the adaptive weights by fusing the magnitude from each sub-network with the top $K$ logits of each group through the designed \textit{fusion module}. Finally, the data-aware classifier is given the final output. Our learning objective and the detail of the fusion module will be described as follows.

For each batch data in practice, we split the samples into three mutual exclusive groups based on their labels. We denote a batch input as $X = \{z_i, c_i\}, i \in \{1, \dots, n\}$, where $z_i$ is the input feature after encoder and $c_i$ is the ground-truth label. Meanwhile, each sample can be further categorized into three mutual exclusive groups with group label $g \in \{1, 2, 3\}$. We try to capture the peculiarity of each sample and find the group label it belongs to. To realize this, we design three sub-networks $\mathcal Q_1, \mathcal Q_2, \mathcal Q_3$ which parameterized 
with $\theta_1$, $\theta_2$ and $\theta_3$ respectively, and a learning objective. Specifically, the learning objective can be formulated as:

\begin{equation}
    r(z_i)=\lambda \left\{
    \begin{array}{ll}
    \max(\xi-\mathcal{Q}_g(z_i; \theta_g), 0)^2 & \text{if}~c_i \in \mathcal G_g, \\
        \max(\mathcal{Q}_g(z_i; \theta_g)-\mu, 0)^2 & \text{otherwise}. \\
    \end{array}
    \right.
\end{equation}

The learning objective aims to regularize the magnitude $\mathcal Q_g(z_i; \theta_g)$, which is the magnitude of the output of sub-network $\mathcal Q_g$, to be higher than $\xi$ for a \textit{positive class}, otherwise lower than $\mu$ for a \textit{negative classes}. For each sub-network $\mathcal Q_g$, classes in the $\mathcal G_g$ are positive classes, other are negative classes. For example, sub-network $\mathcal Q_1$ treats samples in $\mathcal G_1$ as positive and the rest as negative. Intuitively, each sub-network focuses on one group and tries to acquire the difference between groups. To elaborate the construction of the learning objective, we take sub-network $\mathcal{G}_1$ as an example. We leverage its output latent vector and compute the magnitude (L2-norm). We denote the magnitude as $\mathcal Q_1(z_i; \theta_1)$. For notation convenience, we first split $X$ in each group $g$ into positive part $X_g^+$ and negative part $X_g^-$ . Formally, the positive part $X_g^+$ and negative part $X_g^-$ can be expressed as:

\begin{equation}
    \left\{
    \begin{array}{c}
    X_g^+ = \{(z_i, c_i) ~|~ c_i \in \mathcal G_g, (z_i, c_i) \in X\}
    \\
    X_g^- = X \backslash X_g^+
    \end{array}
    \right .
\end{equation}

From another perspective, the $X_g^+$ and $X_g^-$ can be also seen as ID (in-distribution) and OOD (out-of-distribution) samples for sub-network $\mathcal Q_g$ respectively. Based on $X_g^+$ and $X_g^-$, we formulate the regularization objective as:

\begin{equation}
    \label{pos_and_neg_loss}
    \left\{
    \begin{array}{c}
        \mathcal L_{X_g^+} =\frac{1}{\left| X_g^+ \right|} \sum_{z \in X_g^+} \max(\xi-\mathcal Q_g(z; \theta_g), 0)^2; 
        \\
        \mathcal L_{X_g^-}=\frac{1}{\left| X_g^- \right|} \sum_{z \in X_g^-} \max(\mathcal Q_g(z; \theta_g)-\mu, 0)^2
    \end{array}
    \right .
\end{equation}

where $\left| \cdot \right|$ stands for the cardinality in the set. The objective $\mathcal L_{X_g^+}$ is to encourage the output of $\mathcal Q_g$ greater than $\xi$ for positive part $X_g^+$, however, objective $\mathcal L_{X_g^-}$ encourages the output lower than $\mu$ for negative part $X_g^-$. Averaging independently for the $\mathcal L_{X_g^+}$ and the $\mathcal L_{X_g^-}$ can equally contribute to the loss in a positive and negative part level. Otherwise, there will be a big difference in quantitative level between the loss of positive part and the negative part, especially when the distribution of the positive and negative part is extremely imbalanced in the batch. Then the regularization objective for sub-network $Q_g$ can be summed up as:

\begin{equation}
    \mathcal L_{\mathcal Q_g} = \mathcal L_{X_g^+} + \mathcal L_{X_g^-}
\end{equation}

Finally, the regularization objective for all three sub-networks can be written as:

\begin{equation}
    \mathcal L_{REO} = \sum_{g \in 1,2,3} \mathcal L_{\mathcal Q_g}
\end{equation}

To get the final output of MEM, we take a fusion module to fuse the $\mathcal Q(z_i) = [\mathcal Q_1(z_i; \theta_1), \mathcal Q_2(z_i; \theta_2), \mathcal Q_3(z_i; \theta_3
)]$ and top K logits of each group, where K is a hyper-parameter. The fusion module can be implemented as a simple multilayer perceptron (MLP). Suppose the logits are  $\ell_i = \{\ell_i^1, \ell_i^2, \dots \ell_i^m \}$ for representation $z_i$, and the logits of group $g$ can be written as:

\begin{equation}
    \ell_{i,g} = \{\ell_i^j | j \in \{1, \dots, n\}, j \in \mathcal G_g\}
\end{equation}

Suppose $\mathcal T_i = [\mathcal T_i^1, \mathcal T_i^2, \mathcal T_i^3]$ where $\mathcal T_i^g, g \in \{1, 2, 3\}$ is the set of top K values in $\ell_{i,g}$. Then the fusion operation can be formulated as:

\begin{equation}
    w_i=\text{MLP}(\text{concat}(\,\mathcal Q(z_i),\, \mathcal T_i)),
    \label{equ:fusion}
\end{equation}

where $w_i \in \mathbb{R}^3$ denotes the adaptive weights that output from the fusion module, and \textit{concat} means the concatenate operation between the parameters.

\subsection{Data-aware classifier}
After harvesting the adaptive weights $w_i$, we construct a data-aware classifier (DAC) by utilizing the group information in a soft-routing manner. Together with the logits $\ell_i = \{\ell_i^1, \ell_i^2, \dots \ell_i^m \}$ of representation $z_i$, we can formulate our data-aware classifier as:

\begin{equation}
    p_i = \text{softmax}(\{\ell_i^j \cdot w_i^{g(j)} \, | \, j \in \{1, \dots, m\}),
\end{equation}

where $g(j) \in \{1,2,3\}$ denotes the mapping function from the class index~$j$ to the corresponding group.

The loss of data-aware classifier is thus: 

\begin{equation}
    L_{DAC} = -\frac{1}{n}\sum_{i=1}^{n} y_i\text{log}(p_i)
\end{equation}

In the proposed individual mutual exclusive modulator, the two supervisions are employed together to achieve a comprehensive learning, and the final loss is written as:

\begin{equation}
    \mathcal{L} = L_{REO} + L_{DAC}
\end{equation}


\section{Experiments}

\subsection{Experimental setup}

\textbf{Datasets.\ } We conduct our experiments on three large-scale datasets (i.e., ImageNet-LT \cite{liu2019large}, Places-LT \cite{liu2019large}, iNaturalist 2018 \cite{van2018inaturalist}). ImageNet-LT and Places-LT are the long-tailed versions of ImageNet \cite{deng2009imagenet} and Places-365 \cite{zhou2017places} respectively. ImageNet-LT has 1,000 classes and contains 115.8k samples, with maximum of 1,280 samples and minimum 5 samples for a category. 
Places-LT contains 184.5K samples from 365 classes, with class samples ranging from 4,980 to 5. The iNaturalist 2018 is a large-scale species dataset collected in the natural world. It contains 437.5K samples for 8,142 classes.

\textbf{Evaluation protocols.\ } Following previous works \cite{liu2019large, kang2019decoupling}, the top-1 accuracy is adopted for evaluation. Moreover, we follow the setting in \cite{liu2019large} to split the dataset into many-shot (with more than 100 samples), medium-shot (with 20$\sim$100 samples), and few-shot (with less than 20 samples) and report the accuracy of each shot. All the results are reported as a percentage.

\textbf{Implementation details.\ } For ImageNet-LT and iNaturalist 2018, we report results based on ResNeXt-50 \cite{xie2017aggregated} and the transformer networks, i.e., Vit-Base \cite{he2022masked}, Vit-Large \cite{he2022masked} and CVit-Base \cite{gao2022convmae}. We apply MAE in~\cite{he2022masked} to pretrain the Vit-Base and CVit-Base for 400 epochs and Vit-Large for 800 epochs, while 100 epochs train from scratch for ResNeXt-50. For Places-LT, we report results based on ResNet-152 \cite{he2016deep} and the transformer networks mentioned above. We start training from the finetuned classification model from \cite{he2022masked} and \cite{gao2022convmae} for transformer network, while the same setting follows~\cite{liu2019large} for ResNet-152 \cite{he2016deep}. If not specified, for all experiments, we use Adamw \cite{loshchilov2017decoupled} with betas=(0.9, 0.95), cosine learning rate schedule \cite{loshchilov2016sgdr} with learning rate warmup to 1.5e-4 for 40 epochs, mask ratio 0.75 and weight decay 0.05, for MAE pre-training. For supervised finetuning, we train it for 100 epochs with 5 epochs warmup, cosine learning rate schedule, layer-wise learning decay following \cite{he2022masked} and weight decay 0.05. For the training of MEM, we follow the training of decoupling methods in~\cite{kang2019decoupling} to finetune 10 epochs while freezing the backbone. The value K in fusion operation~\ref{equ:fusion} is 5 by default. And the details of the sub-network and the choose for hyper-parameters in the regularization objective can be referred in~\ref{sec: ablation}.

\subsection{Results Comparisons}

\begin{table*}[t]
    \centering
    \renewcommand{\arraystretch}{1.05}
    \begin{tabular}{lccccccc}
        \\ \toprule
        Method & Dataset & Many & Medium & Few & All \\
        \midrule
        MiSLAS\cite{zhong2021improving} & \multirow{8}{*}{ImageNet-LT} & 62.0 & 49.1 & 32.8 & 51.4 \\
        Balanced Softmax~\cite{ren2020balanced} & ~ & 64.1 & 48.2 & 33.4 & 52.3 \\
        LADE~\cite{huang2016learning} & ~ & 64.4 & 47.7 & 34.3 & 52.3 \\
        ACE~\cite{cai2021ace} & ~ &  - & - & - & 56.6 \\
        RIDE~\cite{wang2020long} & ~ & 68.2 & 53.8 & 36.0 & 56.9 \\
        PaCo~\cite{cui2021parametric} & ~ & 68.2 & 58.7 & 41.0 & 60.0 \\
        NCL~\cite{li2022nested} & ~ & - & - & - & 60.5 \\
        TADE~\cite{zhang2021test} & ~ & 68.6 & 61.2 & 47.0 & 62.1 \\
        \cmidrule{1-6}
        MEM(ours) & ~ & 76.0 & 62.0 & 43.7 & \textbf{64.9} \\
        \midrule
        LADE~\cite{huang2016learning} &  \multirow{8}{*}{iNaturalist 2018}  & - & - & - & 69.3 \\
        Balanced Softmax~\cite{ren2020balanced} & ~ & - & - & - & 70.6 \\
        MiSLAS~\cite{zhong2021improving} & ~ & - & - & - & 70.7 \\
        RIDE~\cite{wang2020long} &  ~ & 70.9 & 72.4 & 73.1 & 72.6 \\ 
        ACE~\cite{cai2021ace} &  ~ & - & - & - & 72.9 \\
        PaCo~\cite{cui2021parametric} & ~  & 70.3 & 73.2 & 73.6 & 73.2 \\
        NCL~\cite{li2022nested} &  ~  & - & - & - & 74.9 \\
        TADE~ \cite{zhang2021test} & ~ & 78.3 & 77.0 & 76.7 & 77.0 \\
        \cmidrule{1-6}
        MEM (ours) & ~ & 83.6 & 83.0 & 81.3 & \textbf{82.4} \\
        \midrule
        MiSLAS~\cite{zhong2021improving} &  \multirow{6}{*}{Places-LT}  & - & - & - & 38.3 \\
        LADE~\cite{huang2016learning} & ~ & - & - & - & 39.2 \\
        Balanced Softmax~\cite{ren2020balanced} &  ~ & - & - & - & 39.4 \\
        TADE~ \cite{zhang2021test} &  ~ & 40.4 & 43.2 & 36.8 & 40.9 \\
        PaCo~\cite{cui2021parametric} & ~ & 36.1 & 47.9 & 35.3 & 41.2 \\
        NCL~ \cite{li2022nested} & ~ & - & - & - & 41.8 \\
        \cmidrule{1-6}
        MEM (ours) & ~ & 49.1 & 47.4 & 37.4 & \textbf{46.0} \\
        \bottomrule
    \end{tabular}
    \caption{Comparisons with previous works on dataset ImageNet-LT, iNaturalist 2018 and Places-LT. The results show that our proposed method (MEM) outperforms the state-of-the-art method by a large margin.}
    \label{compare with prev works}
\end{table*}

In this section, we make the comparison in two parts. The first part are about comparison with other decoupling methods which are most relevant to ours and the second part are more comprehensive results with previous methods as shown in Table \ref{compare with prev works}. For the first part, we reproduce the corresponding methods for a fair comparison, and show our superior performances on all datasets mentioned above. For the second part, we compare with the state-of-the-art methods that based on convolutional networks. 
\subsubsection{Comparison with other decoupling methods}

\begin{table}[t]
  \centering
  \small
  \renewcommand{\tabcolsep}{1.8pt}
  \renewcommand{\arraystretch}{1.2}
  \begin{tabular}{@{}lcccc@{}}
    \\ \toprule
    Method & ResNeXt-50 & Vit-Base & Vit-Large & CVit-Base \\
    \midrule
    CE & 45.17 & 53.48 & 59.67 & 57.54 \\
    $\tau$-normalized \cite{kang2019decoupling} & 48.99 & 57.07 & 62.74 & 60.56 \\
    cRT \cite{kang2019decoupling} & 49.25 & 58.06 & 64.04 & 61.85 \\
    LWS \cite{kang2019decoupling} & 48.16 & 57.87 & 63.30 & 61.34 \\
    \midrule
    MEM & \textbf{49.60} & \textbf{58.88} & \textbf{64.85} & \textbf{62.51} \\
    \bottomrule
  \end{tabular}
  \caption{The accuracy on ImageNet-LT with Vision Transformer. We compare with the state-of-the-art decoupling methods on ResNeXt-50, Vit-Base, Vit-Large and CVit-Base.}
  \label{imagenet-lt result}
\end{table}

\begin{figure*}[h]
\begin{center}
\includegraphics[width=0.90\linewidth]{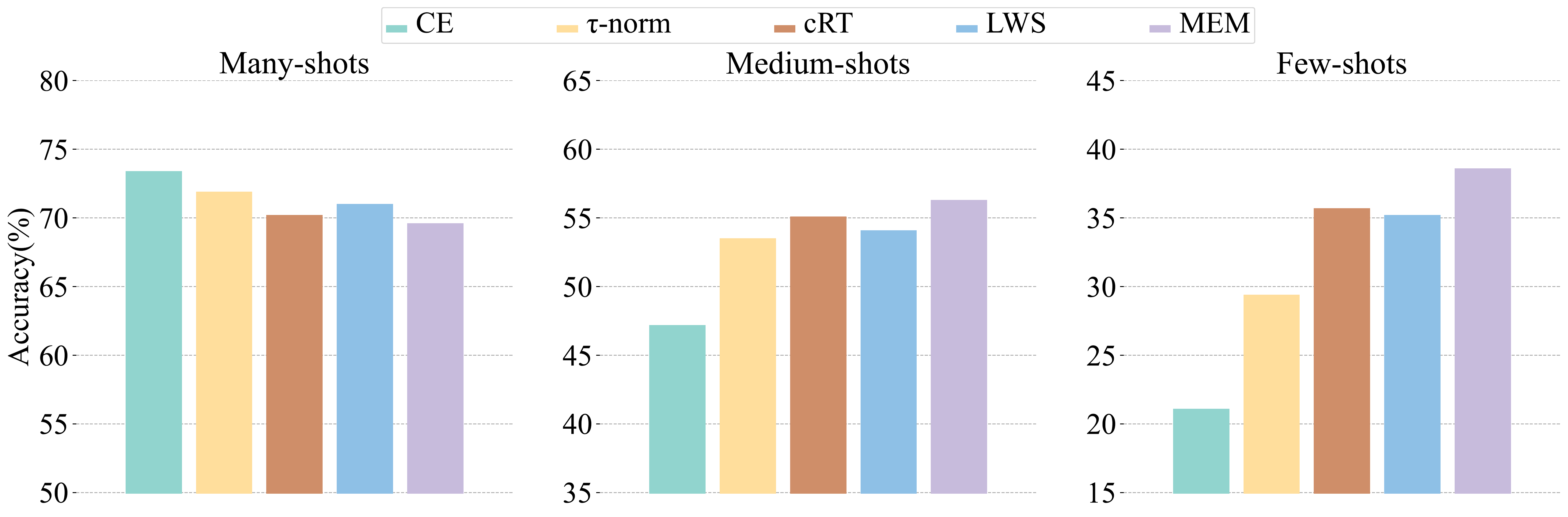}
\end{center}
\caption{An illustration of the performance of each split with Vit-Base on ImageNet-LT. Different colors denote different methods.}
\label{subsets acc}
\end{figure*}

\textbf{ImageNet-LT.\ } Comparisons between different methods on ImageNet-LT are shown in Table \ref{imagenet-lt result}. Most prior works present results based on CNN-based models, but results on recent prevalent structures are lacking. In this paper, we not only provide the comparisons on ResNeXt-50 but also the results on transformer based models. We reproduce the decoupling methods ($\tau$-normalized, cRT and LWS) \cite{kang2019decoupling} on all backbones for a fair comparison. Compare with these methods, without bells and whistles, we achieve a new state-of-the-art for all backbones. Specifically, the MEM surpasses cRT by 0.82\%, 0.81\% and 0.66\% on Vit-Base, Vit-Large and CVit-Base respectively, cRT is the best among all three decouple methods. And we found CVit-Base with fewer parameters but the accuracy is much higher than Vit-Base, e.g., 62.51\% vs. 58.88\%. A reasonable explanation is that the CVit-Base benefits from hybridizing convolutions and transformer blocks. Moreover, as shown in Figure~\ref{subsets acc}, with the proposed mutual exclusive modulator, our method achieves more balance accuracy on each split than other decoupling methods. For example, MEM has higher medium-shot and few-shot accuracy with a slightly lower many-shot accuracy. 

\begin{table}[t]
  \centering
  \small
  \renewcommand{\tabcolsep}{1.8pt}
  \renewcommand{\arraystretch}{1.2}
  \begin{tabular}{@{}lcccc@{}}
    \\ \toprule
    Method & ResNeXt-50 & Vit-Base & Vit-Large & CVit-Base \\
    \midrule
    CE & 60.14 & 72.59 & 79.47 & 78.18 \\
    $\tau$-normalized \cite{kang2019decoupling} & 61.85 & 76.27 & 81.96 & 81.11 \\
    cRT \cite{kang2019decoupling} & 65.21 & 76.29 & 82.09 & 81.14 \\
    LWS \cite{kang2019decoupling} & 63.94 & 76.24 & 82.11 & 81.19 \\
    \midrule
    MEM & \textbf{65.58}  & \textbf{76.63} & \textbf{82.39} & \textbf{81.40} \\
    \bottomrule
  \end{tabular}
  \caption{The accuracy on iNaturalist-2018 with Vision Transformer. We compare with the state-of-the-art decoupling methods on ResNeXt-50, Vit-Base, Vit-Large and CVit-Base.}
  \label{inat-2018 result}
\end{table}

\textbf{iNaturalist 2018.\ } Comparisons on iNaturalist-2018 are shown in Table \ref{inat-2018 result}. Under a fair training setting, MEM surpasses $\tau$-normalized, cRT and LWS consistently across all backbones (i.e., ResNeXt-50, Vit-Base, Vit-Large and CVit-Base). For example, the MEM outperforms cRT, which achieves the highest performance among all three decouple methods, on ResNeXt-50 (65.58\% vs. 65.21\%) and Vit-Base (76.63\% vs. 76.29\%). Once again, the CVit-Base with fewer parameters but has much higher accuracy than Vit-Base (81.40\% vs. 76.63\%). Moreover, we achieve a new state-of-the-art of 82.39\% with Vit-Large that is outperforming the previous best result by 5.4\%, see Table \ref{compare with prev works} for detail. 



\begin{table}[t]
  \centering
  \small
  \renewcommand{\tabcolsep}{1.8pt}
  \renewcommand{\arraystretch}{1.2}
  \begin{tabular}{@{}lcccc@{}}
    \\ \toprule
    Method & ResNet-152 & Vit-Base & Vit-Large & CVit-Base \\
    \midrule
    CE & 30.74 & 35.98 & 37.68 & 36.95 \\
    $\tau$-normalized \cite{kang2019decoupling} & 31.18 & 36.12 & 37.83 & 37.23 \\
    cRT \cite{kang2019decoupling} & 37.16 & 43.48 & 44.81 & 44.11 \\
    LWS \cite{kang2019decoupling} & 36.73 & 42.84 & 44.80 & 43.47 \\
    \midrule
    MEM & \textbf{37.49} & \textbf{44.21} & \textbf{46.04} & \textbf{44.35} \\
    \bottomrule
  \end{tabular}
  \caption{The accuracy on Places-LT with Vision Transformer. We compare with the state-of-the-art decoupling methods on ResNet-152, Vit-Base, Vit-Large and CVit-Base.}
  \label{places result}
  \vspace*{10pt}
\end{table}

\textbf{Places-LT.\ } We further evaluate our MEM on Places-LT dataset. We follow the protocol of \cite{liu2019large}, initialize from a full ImageNet pre-trained model. We present the results after 30 epochs of  fine-tuning, as shown in Table~\ref{places result}. Our MEM exceeds all other approaches, including CE, $\tau$-normalized, cRT, and LWS. Moreover, we achieve the highest accuracy 46.04\% on Vit-Large, which is 4.2\% higher than the best results of previous works, see Table \ref{compare with prev works} for detail.

\subsubsection{More comprehensive comparison}
Here we make a more comprehensive comparison with the previous works on ImageNet-LT, iNaturalist 2018 and Places-LT, as shown in \ref{compare with prev works}. The results of ours are based on the Vit-Large for all datasets. More specifically, our result on ImageNet-LT is 2.8\% higher than the best, e.g., TADE \cite{zhang2021test} which is based on ResNeXt-152~\cite{xie2017aggregated}. On iNaturalist 2018, our result is 5.4\% higher than TADE with ResNeXt-152. Moreover, for Places-LT, our MEM outperforms the best, i.e., NCL~\cite{li2022nested}, by 4.2\%. NCL is based on an ensemble of ResNeXt-50 models. By the comparison with previous CNN based works in Table~\ref{compare with prev works} shows transformer networks could be a better choice for long-tailed recognition.


\subsection{Ablation Study}
\label{sec: ablation}

\textbf{Pre-training for vision transformer.\ } Ablations are conducted on ImageNet-LT with Vit-Base. As shown in Table \ref{pretraining}, Vit-Base without MAE pre-training gets a very low accuracy, i.e., 27.72\% for 100 epochs. It only attains 39.88\% in accuracy when increasing the number of epoch form 100 to 500. However, with the training schedule that 400 epochs for MAE pre-training and 100 epochs for supervised fine-tuning, the accuracy is significantly improved from 39.9\% up to 53.5\%, which shows the necessity of MAE pre-training for transformer based models on long-tailed datasets. One explanation could be that there are insufficient samples in the long-tailed datasets for transformer based models to give play to its strength. From another point of view, the MAE could be regarded as providing with a fair initialization for the following supervised training.

\begin{table*}[t]
    \centering
    \renewcommand{\arraystretch}{1.2}
    \begin{tabular}{ccccccc}
    \\ \toprule
    case & Partitioning strategy & epochs & Many & Medium & Few & All \\
    \midrule
    w/o pre-training & / & 100 & 46.5 & 19.5 & 4.6 & 27.72 \\
    w/o pre-training & / & 500 & 61.0 & 31.6 & 10.9 & 39.88 \\
    w/o pre-training & / & 1000 & 59.5 & 30.0 & 11.2 & 38.59 \\
    w/ pre-training & / & 400+100 & 73.4 & 47.2 & 21.1 & 53.48 \\
    \midrule
    +MEM & Strategy (1) & +10 & 67.9 & 55.9 & 43.0 & 58.67 \\
    +MEM & Strategy (2) & +10 & 71.2 & 55.0 & 34.1 & 58.18 \\
    +MEM & Strategy (3) & +10 & 67.5 & 56.6 & 41.8 & 58.66 \\
    \bottomrule
    \end{tabular}
    \caption{Comparisons on pre-training and without pre-training on ImageNet-LT with Vit-Base. We present the results of without pre-training with supervised learning for 100, 500 and 1000 epochs, respectively. And the result of 400 epochs MAE pre-training and 100 epochs supervise learning is given for comparison. We also present the results of our MEM with different partitioning strategies.}
    \label{pretraining}
    \vspace*{10pt}
\end{table*}


\textbf{Strategies for group partitioning.} An intuitive way for group partition is based on the cardinality of each class. In long-tailed recognition, the classes are divided into \{many, medium, few\}-shot by default and we denote it as strategy (1) here. However, we try to explore whether there are more proper partition methods for mutual exclusive modulator learning. As shown in Table \ref{pretraining}, we first conduct a random and even partition which is denoted as strategy (2). Afterwards, we split the classes evenly according to the cardinality of classes and denote it as strategy (3). We find that even in a random partition, our MEM can work well. But the strategies which utilize the information of the cardinality of each class, i.e., strategy (1) and strategy (3),  work better.


\begin{figure*}
    \centering
    \begin{subfigure}{0.23\linewidth}
    \centering
    \includegraphics[width=1.0\linewidth]{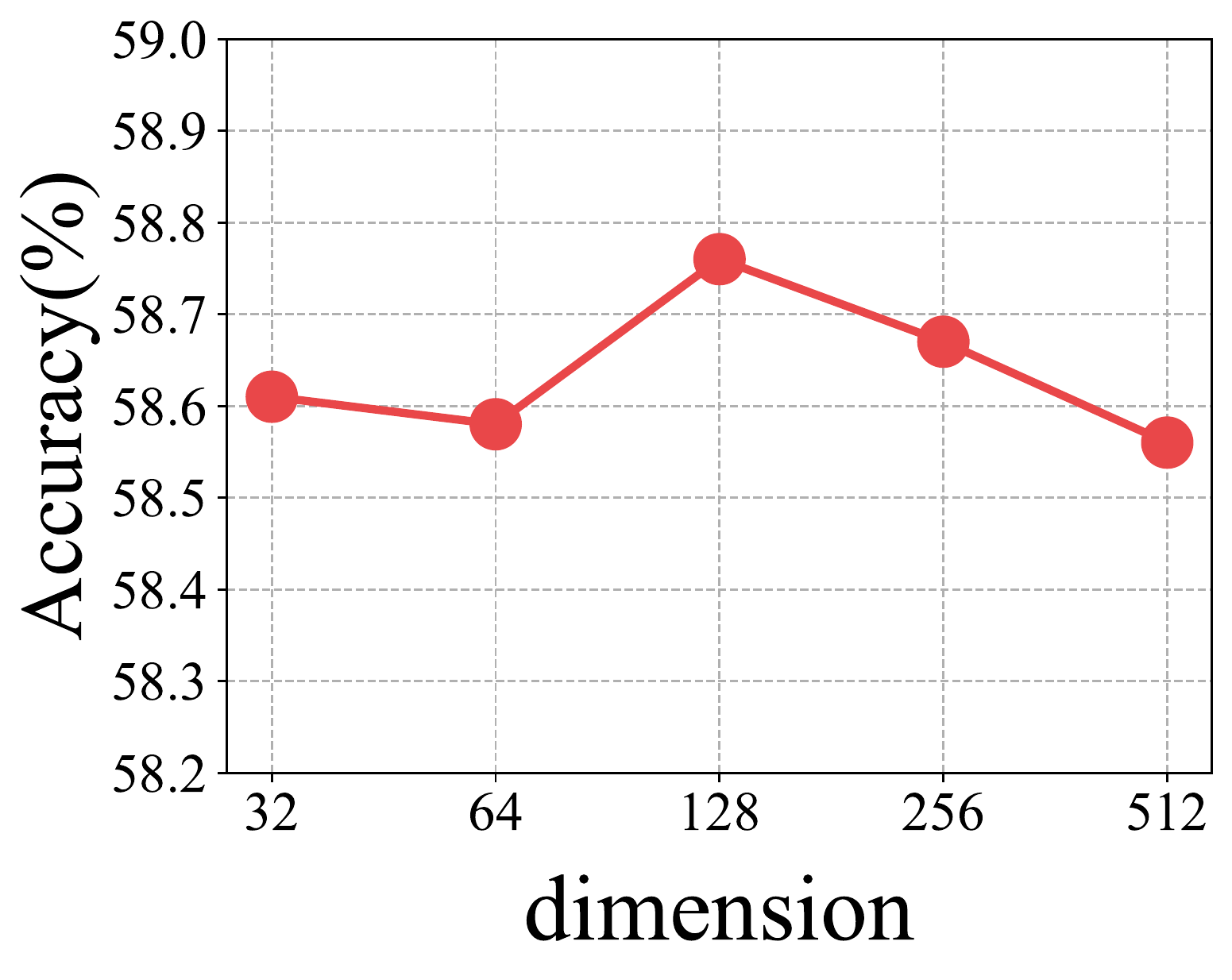}
    \caption{Sub-network width.}
    \label{subnet width}
    \end{subfigure}
    \hfill
    \begin{subfigure}{0.23\linewidth}
    \includegraphics[width=1.0\linewidth]{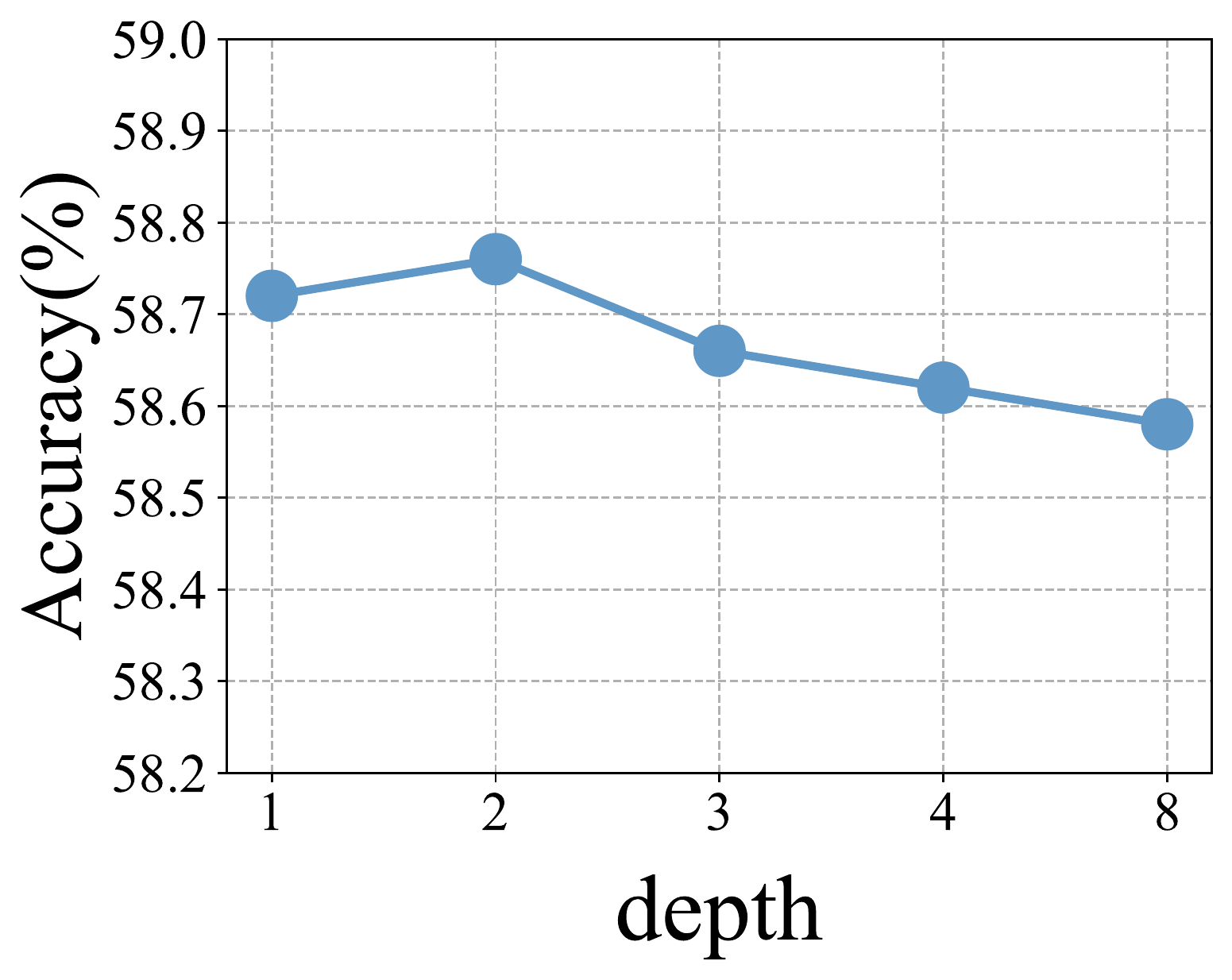}
    \caption{Sub-network depth.}
    \label{subnet depth}
    \end{subfigure}
    \hfill
    \begin{subfigure}{0.23\linewidth}
    \includegraphics[width=1.0\linewidth]{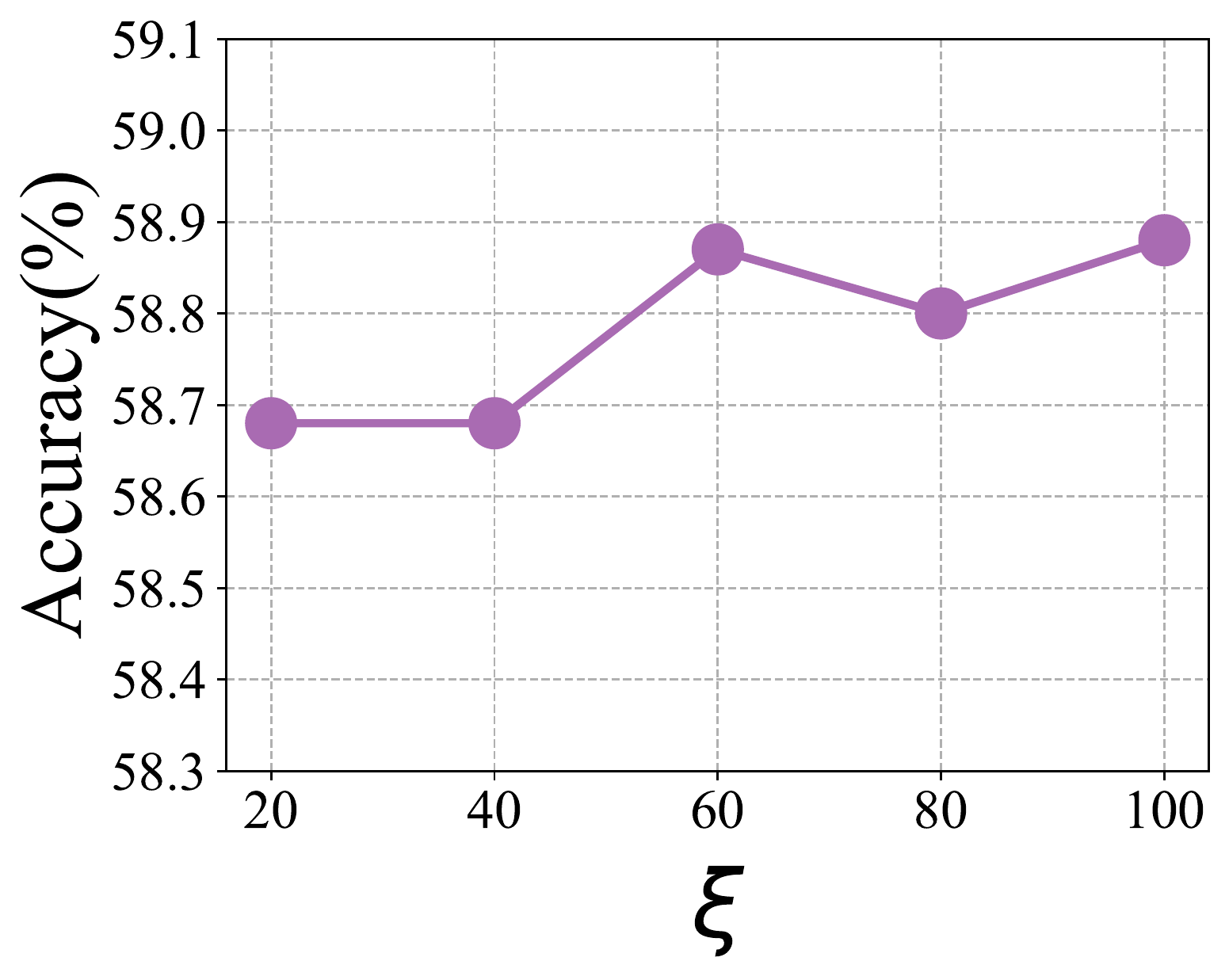}
    \caption{Hyper-parameter $\xi$.}
    \label{ablation xi}
    \end{subfigure}
    \hfill
    \begin{subfigure}{0.23\linewidth}
    \includegraphics[width=1.0\linewidth]{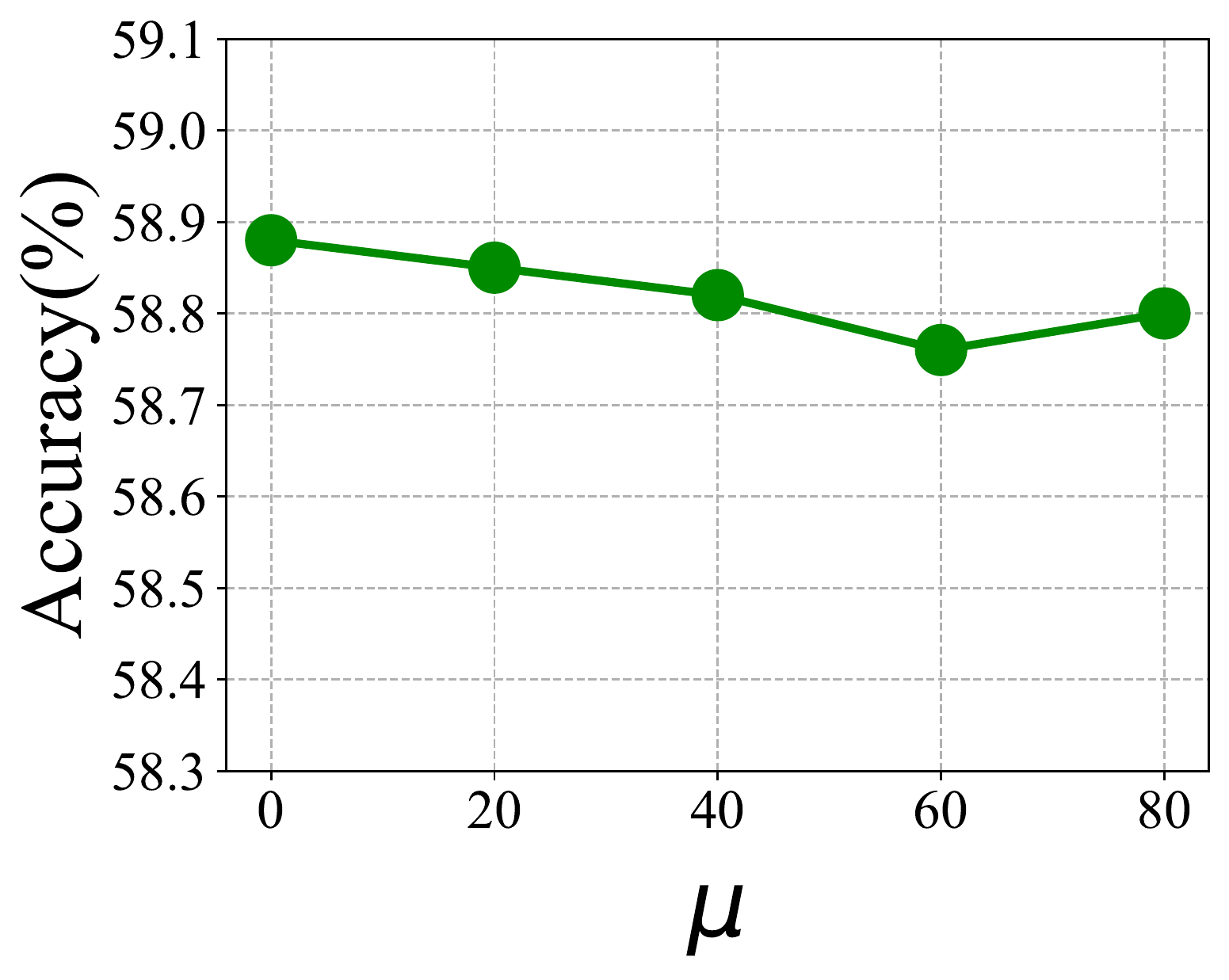}
    \caption{Hyper-parameter $\mu$.}
    \label{ablation mu}
    \end{subfigure}
    \caption{The ablation experiments with Vit-Base on ImageNet-LT. (a) and (b) are the ablation of the capacity of sub-network inside MEM. (c) and (d) are the ablation on the hyper-parameters which used for our regularization objective (see Equation \ref{pos_and_neg_loss}). }
    \label{fig:ablation}
    \vspace*{10pt}
\end{figure*}



\textbf{The capacity of sub-network. } Our sub-network inside MEM can be flexibly designed, we explore from two aspects (i.e., sub-network width and depth), as shown in Figure \ref{subnet width} and \ref{subnet depth}. Figure \ref{subnet width} varies the width (the dimension of embedding) of sub-network, 
while we fix the sub-network depth. And Figure \ref{subnet depth} varies the depth (number of transformer block) of sub-network, while we fix the sub-network width. We found that our sub-network is not sensitive to the width or depth. Interestingly, our sub-network with a single transformer block can yield strong performance (58.72\%). 


\textbf{The hyper-parameters in regularization objective. } We compare the different $\xi$ and $\mu$ for our regularization objective (see Equation \ref{pos_and_neg_loss}), as shown in Figure \ref{ablation xi} and \ref{ablation mu}. Figure \ref{ablation xi} varies the parameter $\xi$ and fix the parameter $\mu$. When $\mu$ is 0, the accuracy has 0.2\% improvement when $\xi$ changes from 20 to 100. This can be explained as the regularization objective needs a large margin between positive classes and negative classes to optimize. And for Figure \ref{ablation mu}, we vary the parameter $\mu$ and under a certain $\xi$. Figure \ref{ablation mu} shows that the accuracy has minor adjustments under different $\mu$. In general, our regularization objective is not sensitive to the hyper-parameters $\xi$ and $\mu$.

\section{Conclusions}
In this paper, we first focus on the behaviors of existing models on three separate groups, \ie, Many, Medium, and Few, in the long-tailed datasets and reveal that the reason for the poor performance is the severe confusion between groups. The model tends to categorize samples in one group to another. Then, we investigate an ideal case that the images are first classified to the right group before the final label is 
predicted. The overall performance has seen huge promotion with this simple assumption. Motivated by this, we thus propose a straightforward structure, which is called Mutual Exclusive Modulator (MEM), to capture the characteristics of each group and the discrepancy among them. It perceives a set of adaptive weights for different groups. Together with original logits, the data-aware classifier make the final prediction by following a soft-routing manner. MEM achieves significantly better results on different backbones, including convolutional networks and transformer networks, compared with the state-of-the-art methods.

{\small
\bibliographystyle{ieee_fullname}
\bibliography{egbib}
}

\end{document}